\documentclass[conference]{IEEEtran}

\usepackage{cite}
\usepackage{amsmath,amssymb,amsfonts}
\usepackage{graphicx}
\usepackage{textcomp}
\usepackage{xcolor}
\usepackage{booktabs}
\usepackage{multirow}
\usepackage{url}
\usepackage{hyperref}
\usepackage{pifont}      
\usepackage{array}
\usepackage{makecell}
\usepackage{algorithm}
\usepackage{algpseudocode}
\usepackage{subcaption}
\newcommand{\cmark}{\textcolor{green!60!black}{\ding{51}}}
\newcommand{\xmark}{\textcolor{red}{\ding{55}}}
\usepackage{amsthm}

\theoremstyle{definition}
\newtheorem{definition}{Definition}
\newtheorem{remark}{Remark}

\begin{document}

\title{FedCARE: A Multi-Objective Personalised Federated Learning Framework for Smart Healthcare}

\author{

\IEEEauthorblockN{
Rojalini Tripathy$^{1,2}$,
Padmalochan Bera$^{2}$,
Shreya Ghosh$^{2}$,
Rajkumar Buyya$^{1}$
}

\IEEEauthorblockA{
$^{1}$Quantum Cloud Computing and Distributed Systems (qCLOUDS) Laboratory\\
School of Computing and Information Systems\\
The University of Melbourne, Parkville, Australia
}

\IEEEauthorblockA{
$^{2}$Department of Computer Science\\
School of Electrical and Computer Sciences\\
Indian Institute of Technology Bhubaneswar, India
}

}

\maketitle


\begin{abstract}

Federated Learning (FL) enables collaborative model training across distributed healthcare institutions without centralising sensitive patient data. However, real-world healthcare federations are often characterised not only by non-IID data, but also by heterogeneous clinical objectives and partially overlapping feature spaces. Different hospitals may optimise distinct and potentially conflicting objectives, such as mortality risk prediction, readmission reduction, or length-of-stay estimation, while also retaining institution-specific clinical features that cannot be shared with other participants. Existing personalised FL methods mainly address statistical heterogeneity, whereas multi-objective FL approaches typically learn a shared global model without explicit client-level adaptation. To address these limitations, we propose \textbf{FedCARE}, a multi-objective personalised FL framework for smart healthcare services. FedCARE follows a two-stage training strategy. First, it learns a shared global backbone from common clinical features using Pareto-driven multi-objective federated optimisation. Second, each client independently fine-tunes the shared backbone using its private features and local clinical objectives, enabling institution-specific personalisation without additional communication overhead. We implement FedCARE in a cloud-based client-server federated deployment on the Melbourne Research Cloud and evaluate it on two real-world healthcare datasets, MIMIC-III and Diabetes 130-US Hospitals. Experimental results show that FedCARE consistently outperforms standard FL, multi-objective FL, and personalised FL baselines, achieving up to 12.5\% AUROC improvement and 32.0\% MAE reduction over FedAvg.

\end{abstract}



\section{Introduction}

Training Machine Learning (ML) models on Electronic Health Records (EHRs) collected across multiple healthcare organisations is essential for building intelligent and scalable clinical decision-support systems. However, centralising patient records is often infeasible due to privacy regulations such as HIPAA \cite{act1996health} and GDPR \cite{GDPR_2016}, as well as institutional concerns regarding data ownership, governance, and regulatory compliance. Federated Learning (FL) \cite{mcmahan2017communication} has emerged as a promising paradigm for distributed healthcare intelligence by enabling collaborative model training without requiring raw patient data to leave local institutions. Therefore, FL has been widely adopted in smart healthcare applications, including disease prediction, patient monitoring, clinical risk assessment, and intelligent healthcare service delivery \cite{nguyen2022federated, annappa2024fedcure, singh2023energy}.

\begin{figure}[!t]
	\centering
	\includegraphics[scale=0.33]{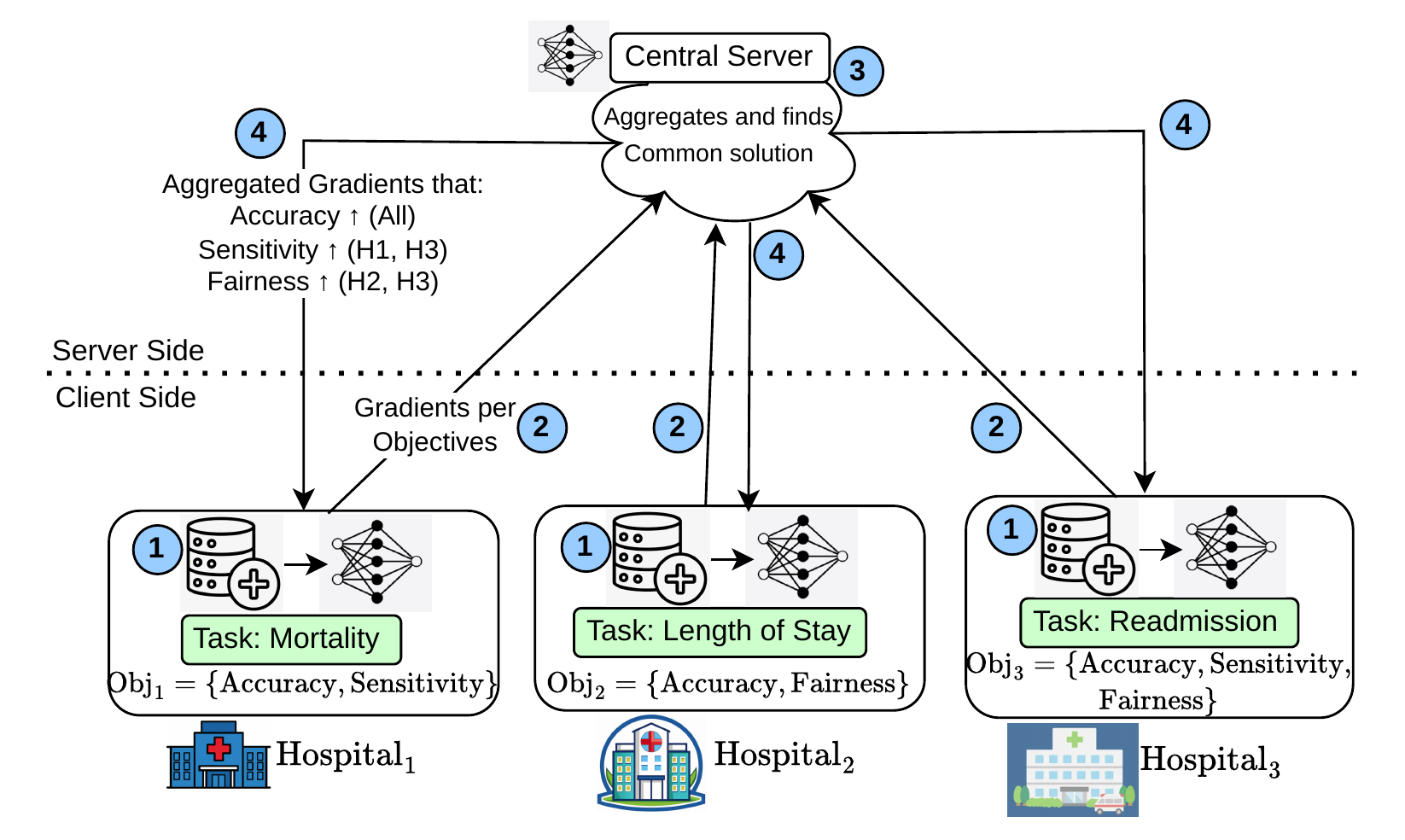}
	\caption{A multi-objective federated healthcare scenario where hospitals optimise different subsets of clinical objectives such as accuracy, fairness, and sensitivity.}
  \label{fig:1}
\end{figure}

Despite its promise, real-world healthcare federated environments are highly heterogeneous \cite{chen2025advances}. Existing studies primarily focus on \textit{data heterogeneity}, where client data distributions differ; \textit{system heterogeneity}, where clients have different computational and communication capabilities; and \textit{model heterogeneity}, where clients use different architectures or optimisation settings \cite{hartmann2025fedpref}. However, healthcare federations also exhibit \textit{objective heterogeneity}, where different institutions optimise distinct and potentially conflicting clinical objectives depending on local patient populations, clinical priorities, resource availability, and risk tolerance.

In practical deployments, hospitals and clinical units rarely optimise identical objectives. For example, a tertiary ICU may prioritise high sensitivity in mortality prediction because missing a high-risk patient can be clinically critical, whereas a resource-constrained hospital may prioritise accurate length-of-stay estimation to improve bed allocation and discharge planning. Similarly, another institution may focus on reducing readmission risk to support post-discharge care management. Figure~\ref{fig:1} illustrates such a multi-objective federated healthcare scenario, where hospitals collaborate while pursuing different clinical objectives. These objectives can induce conflicting gradient directions during federated aggregation. Standard FL methods based on uniform averaging or single-objective optimisation are not designed to resolve such conflicts, which can degrade global model quality and limit personalisation.

In addition to objective heterogeneity, healthcare federated systems often suffer from \textit{feature space heterogeneity}. Existing FL literature mainly studies quantity skew, label distribution skew, feature distribution skew, and concept drift \cite{ye2023heterogeneous, li2022federated}. However, healthcare institutions may collect data using different clinical workflows, EHR schemas, sensor configurations, and diagnostic procedures. As a result, clients may share only a subset of common clinical attributes while retaining institution-specific private features unavailable to other participants. Vertical Federated Learning (VFL) \cite{liu2024vertical} requires reliable sample alignment across institutions, which is often impractical due to privacy constraints and inconsistent patient identifiers. Transfer learning approaches \cite{zhang2021parameterized} may also suffer from negative transfer when institutional feature spaces, distributions, and objectives differ substantially.

Although personalised FL and multi-objective FL have been studied independently, existing methods remain insufficient for such healthcare settings. Personalised FL methods mainly address statistical heterogeneity while generally assuming identical optimisation objectives across clients. In contrast, multi-objective FL methods optimise multiple objectives globally but typically rely on a shared model without explicit institution-specific adaptation. Therefore, neither class of methods can jointly address conflicting clinical objectives and partially overlapping feature spaces. From a service computing perspective, this creates a critical challenge: healthcare federated systems must support privacy-preserving collaboration, scalable distributed optimisation, and institution-specific service adaptation under heterogeneous operational and clinical constraints.

To address this challenge, we propose \textbf{FedCARE}, a multi-objective personalised FL framework for heterogeneous healthcare environments. FedCARE jointly addresses \textit{objective heterogeneity} and \textit{feature space disparity} through a two-stage learning paradigm. In the first stage, clients collaboratively learn a shared global backbone using common clinical features through Pareto-driven multi-objective federated optimisation. This stage aims to obtain a Pareto-stationary shared representation that balances conflicting objectives across distributed healthcare clients. In the second stage, each institution independently fine-tunes the shared backbone using its private feature space and local clinical objectives, enabling personalised adaptation without additional communication overhead. This design preserves global knowledge sharing while supporting institution-specific specialisation and scalable personalised healthcare services. We implement FedCARE in a cloud-based client-server federated deployment on the Melbourne Research Cloud (MRC) and evaluate it on two real-world healthcare datasets: MIMIC-III \cite{johnson2016mimic} and Diabetes 130-US Hospitals \cite{strack2014impact}. We compare FedCARE with standard FL methods, including FedAvg \cite{mcmahan2017communication} and FedProx \cite{li2020federated}, multi-objective FL methods including FMGDA and FSMGDA \cite{yang2023federated}, and a personalised healthcare FL framework, PAFNet \cite{11170414}. The main contributions of this paper are summarised as follows:

\begin{itemize}

\item We identify \textit{objective heterogeneity} and \textit{feature space disparity} as two important yet jointly underexplored challenges in federated healthcare systems, where institutions may optimise different clinical objectives while sharing only partially overlapping feature spaces.

\item We propose \textbf{FedCARE}, a two-stage Pareto-driven personalised FL framework that decouples shared multi-objective representation learning from institution-specific adaptation.

\item We implement FedCARE in a cloud-based client-server federated healthcare environment on the MRC, demonstrating its practical feasibility for distributed healthcare service computing.

\item We conduct extensive experiments on MIMIC-III and Diabetes 130-US Hospitals datasets, showing that FedCARE consistently outperforms standard FL, multi-objective FL, and personalised FL baselines, achieving up to 12.5\% AUROC improvement and 32.0\% MAE reduction.

\end{itemize}

\section{Related Work}
In this section, we review existing research in four major directions relevant to our work: personalised healthcare using FL \cite{11170414, annappa2024fedcure, barros2024novel, wu2020fedhome, howlader2026federated}, multi-objective optimisation in FL \cite{yang2023federated, hartmann2023mofl, zhu2019multi, hartmann2025fedpref}, feature-space heterogeneity in healthcare, and federated multi-task learning \cite{sattler2020clustered, ghosh2020efficient, skovajsova2025review}.
\subsection{personalised Healthcare using FL}
\par FL has been widely adopted in personalised healthcare \cite{11170414,annappa2024fedcure, howlader2026federated, nguyen2022federated, chaddad2023federated}, improving scalability and data privacy in collaborative training. Molaei et al. \cite{11170414} proposed a Personalised Attention-based Federated Graph Network (PAFNet) to address data-view heterogeneity in EHRs. Instead of manually aligning features across clients, PAFNet learns client-specific projection layers to map heterogeneous local feature spaces into a common latent space. It uses trainable parameter masks to personalise the global model for each client and employs a peer-attentive aggregation mechanism that assigns higher weights to functionally similar clients. Annappa et al. \cite{annappa2024fedcure} proposed FedCure, a personalised FL framework for intelligent IoMT-based healthcare applications. FedCure addresses data, device, and model heterogeneity through computational offloading to edge servers and multiple personalisation strategies, including hypernetwork-based, meta-learning-based, regularisation-based, and knowledge distillation-based approaches. However, these frameworks assume a single shared objective across all participating clients. Recently, Howlader et al. \cite{howlader2026federated} presented a survey of FL applications in healthcare informatics. The authors examined five core challenges in healthcare FL: data heterogeneity, communication heterogeneity, computational heterogeneity, privacy, and security. However, the survey does not discuss multi-objective heterogeneity in healthcare, thereby highlighting an important research gap and motivating the proposed work. 
\subsection{Multi-Objective Optimization in FL}
\par In 2023, two studies introduced the concept of multi-objective optimization in FL \cite{yang2023federated, hartmann2023mofl}. In \cite{yang2023federated} Yang et al. propose two federated multi-objective optimization algorithms: Federated Multi-Gradient Descent Averaging (FMGDA), which employs full local gradients, and Federated Stochastic Multi-Gradient Descent Averaging (FSMGDA), which uses stochastic gradients for scalability. They prove that both algorithms converge to a Pareto-stationary solution. However, the server maintains and broadcasts a single shared global model to all clients, with no personalization for potentially conflicting objective distributions across clients. Furthermore, while the theoretical analysis establishes convergence guarantees, the impact of objective heterogeneity on solution quality is not examined in further empirical detail. In \cite{hartmann2023mofl} Hartmann et al. propose MOFL/D, a federated multi-objective learning framework based on decomposition, targeting problems where user preferences are unknown at training time. The framework decomposes the multi-objective problem into scalarised single-objective subproblems via linear weight vectors. Each subproblem is solved using a full federated training run, maintaining a growing Pareto front. While the framework is modular and produces a diverse set of Pareto-optimal solutions, it assumes that all clients share the same set of objectives. Furthermore, each subproblem requires an independent federated training run, resulting in computational overhead that limits scalability. Prior to this, \cite{zhu2019multi}, Zhu and Jin propose a multi-objective FL framework that applies multi-objective optimisation as a meta-level tool to FL itself, rather than treating the learning problem as inherently multi-objective. This approach is not closely aligned with our research direction.
Recently, Hartmann et al. \cite{hartmann2025fedpref} proposed FedPref, in which each client has distinct preferences. All clients share the same set of objectives but assign different importance weights to each objective. FedPref uses a personalised FL algorithm that combines recursive spectral clustering with adaptive weighted aggregation. However, the framework transforms the multi-objective problem into a single-objective optimization, where each client aggregates multiple objectives into a weighted sum based on its preference weights and learns a personalised model accordingly. 
\subsection{Federated Multi-Task Learning (FMTL)}
\par Another research direction evaluates FL frameworks for handling multiple related tasks by clustering clients with similar characteristics, known as FMTL \cite{sattler2020clustered, ghosh2020efficient}. FMTL  addresses data heterogeneity by grouping clients with similar data distributions, where each cluster trains a separate model tailored to its local data. Clusters are generally formed based on the similarity of client gradient updates, enabling effective knowledge sharing while reducing the impact of incompatible data \cite{skovajsova2025review}. To the best of our knowledge,  there is no existing research on multi-objective FL in personalised healthcare, where each healthcare organisation has a different objective function and the prediction task may be the same or different.

Table~\ref{tab:feature_comparison} presents a qualitative comparison between existing multi-objective FL approaches and the proposed framework across five key features. Existing multi-objective FL methods~\cite{yang2023federated, hartmann2023mofl, hartmann2025fedpref} do not address model personalization arising from feature heterogeneity across clients. Furthermore, some works~\cite{hartmann2023mofl, hartmann2025fedpref} assume that all clients share the same set of objectives, which does not capture real-world objective heterogeneity. Healthcare environments are inherently heterogeneous in terms of data distributions, patient populations, and clinical objectives, yet multi-objective FL has not been applied for healthcare applications. The proposed FedCARE framework addresses multi-objective optimization, feature heterogeneity, and model personalization in healthcare applications.

\begin{table}[t]
\centering
\caption{Qualitative feature comparison of related works and our proposed framework.}
\label{tab:feature_comparison}
\resizebox{\columnwidth}{!}{%
\begin{tabular}{l c c c c c c}
\toprule
Feature & \cite{yang2023federated} &
\cite{hartmann2023mofl} & \cite{hartmann2025fedpref} &
\cite{11170414} & \cite{annappa2024fedcure} & Ours \\
\midrule
Multi Objective & \cmark & \cmark & \cmark & \xmark& \xmark & \cmark \\
Data Heterogeneity & \xmark & \xmark & \xmark & \cmark & \cmark & \cmark \\
Personalisation & \xmark & \xmark & \cmark& \cmark & \cmark & \cmark \\
Heterogeneous Objective Sets& \cmark& \xmark & \xmark& \xmark & \xmark & \cmark \\
Healthcare Application & \xmark & \xmark & \xmark & \cmark & \cmark & \cmark \\
\bottomrule
\end{tabular}}
\end{table}

\section{Preliminaries}
\subsection{Multi-Objective Federated Learning}
\noindent
Many real-world learning problems require optimising multiple objectives simultaneously instead of combining them into a single objective. In this work, a clinical task refers to the prediction problem, such as mortality, readmission, or length-of-stay prediction. An optimisation objective refers to the loss or clinical preference optimised for a task, such as prediction loss, sensitivity-oriented loss, or fairness-aware loss. Evaluation metrics such as AUROC and MAE are used only to assess the learned models. Multi-objective optimisation \cite{momma2022multi, yang2023federated} addresses this by finding a solution $\mathbf{x} \in \mathcal{D} \subseteq \mathbb{R}^d$ that minimises a set of $S$ objective functions:
\begin{equation}
    \min_{\mathbf{x} \in \mathcal{D}} \; \mathbf{F}(\mathbf{x}) 
    \;=\; \bigl[f_1(\mathbf{x}),\, f_2(\mathbf{x}),\, \ldots,\, f_S(\mathbf{x})\bigr]^\top,
    \label{eq:moo}
\end{equation}
where each $f_s : \mathbb{R}^d \to \mathbb{R}$, $s \in [S]$. Because objectives often conflict, a single solution may not minimise all of them at once. Therefore, multi-objective optimisation problems use the concept of Pareto optimality:

\begin{definition}[Pareto Optimality and Pareto Stationarity]
A solution point $\mathbf{x}$ dominates $\mathbf{y}$ if $f_s(\mathbf{x}) \leq f_s(\mathbf{y})$ for all $s \in [S]$ and $f_s(\mathbf{x}) < f_s(\mathbf{y})$ for at least one $s$. 
Since finding a Pareto-optimal solution is generally NP-hard, a practical relaxation is Pareto stationarity. A point $\mathbf{x}$ is Pareto stationary if there 
exists no descent direction $\mathbf{d} \in \mathbb{R}^d$ such that 
$\nabla f_s(\mathbf{x})^\top \mathbf{d} < 0$ for all $s \in [S]$. 
\end{definition}
\noindent
Gradient-based multi-objective algorithms find Pareto-stationary points by identifying a common descent direction that simultaneously reduces all objectives. 
\noindent
In multi-objective FL, different clients may have different subsets 
of the $S$ objectives. Client $k$ participates in task $s$ only if 
$a_{s,k} = 1$ in the binary indicator matrix 
$\mathbf{A} \in \{0,1\}^{S \times K}$. Multi-objective FL 
\cite{yang2023federated} allows $K$ distributed clients to 
collaboratively find a Pareto-stationary solution while keeping their local data private.

\begin{definition}[Multi-Objective Federated Learning (MOFL)]
\label{def:mofl}
Consider a system with $K$ clients and $S$ objectives. Let
$f_{k,s}(\mathbf{x})$ denote the local objective associated with
objective $s$ at client $k$, and let
$\mathbf{A} \in \{0,1\}^{S \times K}$ be a binary indicator matrix,
where $a_{s,k}=1$ if client $k$ participates in objective $s$, and
$a_{s,k}=0$ otherwise. The MOFL problem finds a Pareto-stationary
solution by jointly minimising the vector of global objectives defined in Eq.~(\ref{eq:moo}),
where each global objective is obtained by averaging the corresponding
local objectives over all participating clients:
\begin{equation}
\begin{aligned}
f_s(\mathbf{x})
&=
\frac{1}{|\mathcal{K}_s|}
\sum_{k\in\mathcal{K}_s}
f_{k,s}(\mathbf{x}),\\
\mathcal{K}_s
&=
\{k\in[K]:a_{s,k}=1\},
\end{aligned}
\label{eq:global_obj}
\end{equation}
where $\mathcal{K}_s$ denotes the set of clients participating in
objective $s$.
\end{definition}

\subsection{Feature Space Disparity in Federated Learning}

\noindent
In standard FL, all clients are assumed to operate over the same feature space. In practice, federated healthcare systems collect and maintain distinct sets of clinical variables based on their data-acquisition protocols, medical practices, and available resources. Therefore, clients exhibit feature-space disparity, where only a subset of features is shared across all clients while the remaining features are institution-specific.

\begin{definition}[Feature Partition]
\label{def:feature_partition}
For a system with $K$ clients, let
$\mathcal{X}_c \subseteq \mathbb{R}^{d_c}$ denote the common feature
space shared across all clients, and let
$\mathcal{X}_{u,k} \subseteq \mathbb{R}^{d_k}$ denote the uncommon
feature space available at client $k$. The local data sample at
client $k$ is therefore
\[
\mathbf{x}_k=(\mathbf{x}_c,\mathbf{x}_{u,k})
\in
\mathcal{X}_c\times\mathcal{X}_{u,k}.
\]
\end{definition}

\noindent
We acknowledge that, in addition to feature-space disparity, other forms of heterogeneity may naturally arise in healthcare federated systems and can also be incorporated within the FedCARE framework. We describe below two common types of heterogeneity.

\begin{itemize}
    \item \textbf{Feature Distribution Heterogeneity:} It occurs when the marginal distribution of the common feature space differs across clients. Formally,
    \[
    P_k(\mathbf{x}_c) \neq P_j(\mathbf{x}_c),
    \quad \text{for some } k \neq j.
    \]

    \item \textbf{Feature-Label Relationship Heterogeneity:} It occurs when the conditional distribution of labels given features varies across clients. Formally,
    \[
    P_k(y \mid \mathbf{x}_c,\mathbf{x}_{u,k})
    \neq
    P_j(y \mid \mathbf{x}_c,\mathbf{x}_{u,j}),
    \quad \text{for some } k \neq j.
    \]
    Here, $k,j \in [K]$ denote two distinct clients. This captures situations where the same clinical features may have different relationships with the prediction task across institutions.
\end{itemize}

\begin{remark}[Reduction to MOFL]
\label{rem:reduction}
When $\mathcal{X}_{u,k}=\emptyset$ for all $k\in[K]$, all clients share the same feature space, and the proposed framework reduces to the standard MOFL problem described in Definition~\ref{def:mofl}.
\end{remark}
\section{Proposed FedCARE Framework}

\begin{figure*}[t]
    \centering
    \includegraphics[width=\textwidth]{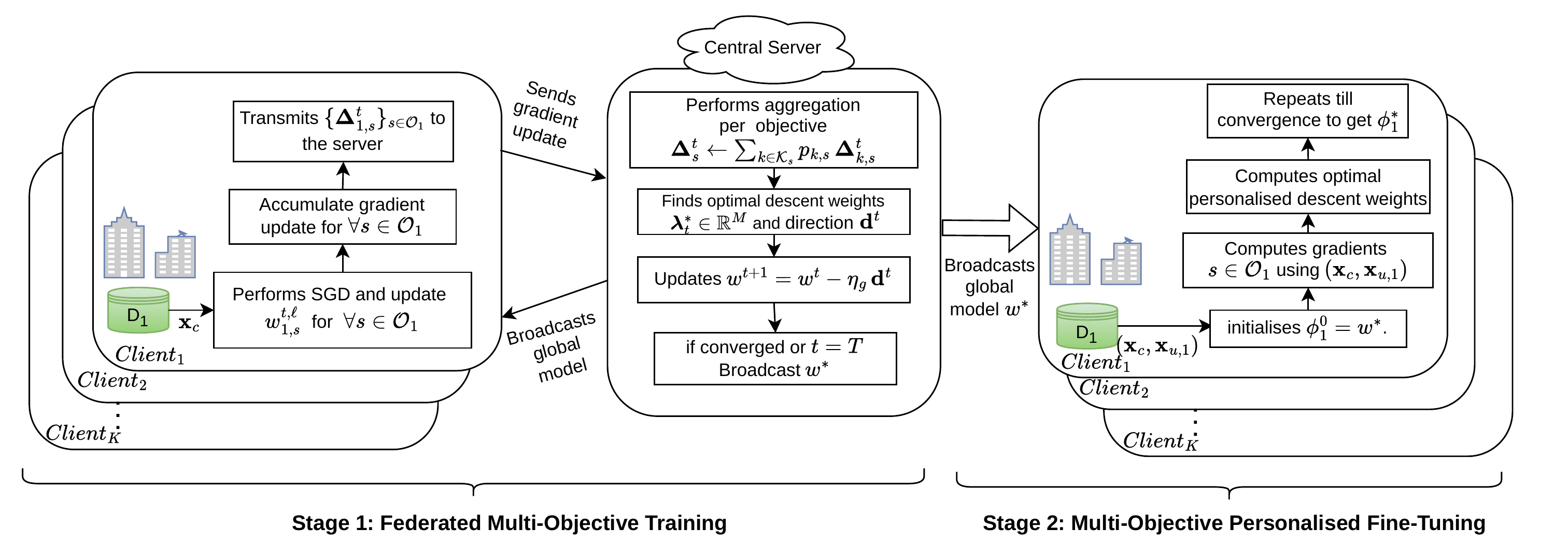}
    \caption{Architecture of FedCARE. Stage~1: clients train on common features, solve a Pareto QP to obtain optimal weights, and broadcast the converged backbone to all clients. Stage~2: each client independently fine-tunes a personalised model using the full feature set.}
    \label{fig:2}
\end{figure*}

The proposed FedCARE framework consists of a two-stage federated training process. In the first stage, the framework collaboratively trains on common data features and multiple objectives across all clients to find a Pareto-stationary solution that satisfies all objectives. In the second stage, each client personalises the global model using its own uncommon features and client-specific objectives. Let's consider $K$ as the total number of clients and $S=\left|\bigcup_{k=1}^{K}\mathcal{O}_k\right|$ as the total number of distinct objectives across all clients, where $m_k=|\mathcal{O}_k|\leq S$ denotes the number of objectives at client $k$. Note that the binary indicator matrix $\mathbf{A}$ from Definition~\ref{def:mofl} is equivalently represented here through the set-based notation
$\mathcal{O}_k$ and $\mathcal{K}_s$ for clarity. Let $\mathcal{K}=\{1,2,\ldots,K\}$ denote the set of $K$ participating clients, where each client $k\in\mathcal{K}$ holds a local dataset:
\begin{equation}
    D_k
    =
    \bigl\{
    (\mathbf{x}_c^{(n)},\,\mathbf{x}_{u,k}^{(n)},\,
    \mathbf{y}_k^{(n)})
    \bigr\}_{n=1}^{N_k},
    \label{eq:local_dataset}
\end{equation}
where $\mathbf{x}_c^{(n)}\in\mathcal{X}_c$ is the common feature
vector, $\mathbf{x}_{u,k}^{(n)}\in\mathcal{X}_{u,k}$ is the
client-specific feature vector, and
$\mathbf{y}_k^{(n)}$ is the label vector corresponding to the
objectives in $\mathcal{O}_k$. Let $N_k=|D_k|$ denote the local
dataset size. Each client $k$ has a set of clinical objectives:
\begin{equation}
    \mathcal{O}_k
    =
    \{\,s\in[S]:a_{s,k}=1\,\},
    \label{eq:local_objectives}
\end{equation}
where each objective index $s\in\mathcal{O}_k$ corresponds to the
local objective function $f_{k,s}(\mathbf{x})$. The architectural workflow of FedCARE is illustrated in
Figure~\ref{fig:2}.
\subsection{Stage 1: Federated Multi-Objective Training}

The objective of Stage~1 is to learn a shared global backbone model $w^*$ that is Pareto-stationary with respect to all $S$ objectives, using the common features $\mathbf{x}_c$ available at every client. This stage follows the MOFL formulation of Definition~\ref{def:mofl} on the common feature space $\mathcal{X}_c$. We perform federated training until convergence or for $T$ communication rounds between the server and the clients. The server initialises the global backbone $w^0$ and broadcasts it to all clients before training. At each communication round, each client $k \in \mathcal{K}$ first downloads the current global backbone $w^t$ broadcast by the server and initialises the local model for each objective $s \in \mathcal{O}_k$:

\begin{equation}
    w^{t,0}_{k,s} \;=\; w^t, \qquad \forall\; s \in \mathcal{O}_k.
    \label{eq:init_local}
\end{equation}
Using the common features $\mathbf{x}_c$ from the local dataset $D_k$, the client performs $\tau$ local stochastic gradient descent steps for each objective. At each local step $\ell \in \{1, \dots, \tau\}$:

\begin{equation}
    w^{t,\ell}_{k,s} \;=\; w^{t,\ell-1}_{k,s} \;-\; \eta_l\,
    \nabla f_{k,s}\!\bigl(w^{t,\ell-1}_{k,s};\, \mathbf{x}_c,\,
    \xi^{t,\ell}_k\bigr),
    \label{eq:local_step}
\end{equation}
where $\eta_l > 0$ is the local learning rate and $\xi^{t,\ell}_k$ is a mini-batch sampled uniformly from $D_k$. The client then computes the accumulated gradient update for each objective $s \in \mathcal{O}_k$:

\begin{equation}
    \boldsymbol{\Delta}^t_{k,s} \;=\; \sum_{\ell=1}^{\tau}
    \nabla f_{k,s}\!\bigl(w^{t,\ell-1}_{k,s};\, \mathbf{x}_c,\,
    \xi^{t,\ell}_k\bigr).
    \label{eq:local_accum}
\end{equation}
Finally, the client transmits $\{\boldsymbol{\Delta}^t_{k,s}\}_{s \in \mathcal{O}_k}$ to the server. After receiving updates from clients, the server aggregates updates
per objective. For each global objective $s \in [S]$,
it aggregates updates from all clients that participate in objective $s$:

\begin{equation}
    \boldsymbol{\Delta}^t_s \;=\; \sum_{k \in \mathcal{K}_s}
    p_{k,s}\; \boldsymbol{\Delta}^t_{k,s},
    \qquad
    \mathcal{K}_s \;=\; \bigl\{k \in [K] : s \in \mathcal{O}_k\bigr\},
    \label{eq:server_agg}
\end{equation}
where $p_{k,s} = N_k / \sum_{j \in \mathcal{K}_s} N_j$ denotes the
normalised weight of client $k$ for objective $s$. The server then solves a quadratic program (QP) to find the optimal descent weights $\boldsymbol{\lambda}^*_t \in \mathbb{R}^{S}$:

\begin{equation}
\begin{aligned}
\boldsymbol{\lambda}^*_t
&= \arg\min_{\boldsymbol{\lambda} \in \mathcal{C}}
\left\| \sum_{s=1}^{S} \lambda_s\, \boldsymbol{\Delta}^t_s \right\|^2, \\
\mathcal{C}
&= \Bigl\{\boldsymbol{\lambda} \in [0,1]^S \;:\;
\sum_{s=1}^{S} \lambda_s = 1 \Bigr\}.
\end{aligned}
\label{eq:server_qp}
\end{equation}
The solution $\boldsymbol{\lambda}^*_t$ assigns higher weights to objectives with better-aligned gradients, producing a balanced common descent direction. The server computes the descent direction as:
\begin{equation}
    \mathbf{d}^t \;=\; \sum_{s=1}^{S} \lambda^*_{t,s}\;
    \boldsymbol{\Delta}^t_s.
    \label{eq:descent_dir}
\end{equation}
The server updates the global model using this descent direction:
\begin{equation}
    w^{t+1} \;=\; w^t \;-\; \eta_g\, \mathbf{d}^t,
    \label{eq:server_update}
\end{equation}
where $\eta_g > 0$ is the global learning rate. Then, the server checks for convergence. If $\|\mathbf{d}^t\|^2 \leq \epsilon$ for a tolerance $\epsilon > 0$, or if $t = T$, it terminates Stage~1. Otherwise, it broadcasts $w^{t+1}$ to all clients and proceeds to round $t+1$. Upon termination, the server broadcasts the converged backbone $w^* \triangleq w^T$ to all clients as the initialisation for Stage~2. The backbone $w^*$ provides a shared representation over the common feature space for subsequent client-specific personalisation. The complete federated multi-objective training procedure for Stage~1 is presented in Algorithm~\ref{alg:stage1}.

\subsection{Stage 2: Multi-Objective Personalised Fine-Tuning}

The goal of Stage~2 is to personalise the shared backbone $w^*$ at each client using its private uncommon features $\mathbf{x}_{u,k}$ and client-specific objectives $\mathcal{O}_k$. We perform this stage independently for each client, without communicating with the server or other clients, thereby preserving privacy and incurring no additional communication cost. However, if clients with similar objective vectors wish to collaborate, FedCARE can be extended to cluster such clients prior to training, enabling collaborative personalization. Each client $k \in \mathcal{K}$ initialises its personalised model from the global backbone:

\begin{equation}
    \phi^0_k \;=\; w^*, \qquad \forall\; k \in \mathcal{K}.
    \label{eq:phase2_init}
\end{equation}
Each client then performs $\tau$ local iterations. At each iteration $\ell$, the client computes the gradient of each local objective $f_{k,s}$, where $s\in\mathcal{O}_k$, with respect to $\phi_k$ using the full feature set $(\mathbf{x}_c,\mathbf{x}_{u,k})$:
\begin{equation}
    \mathbf{g}^{\ell}_{k,s}
    =
    \nabla_{\phi_k}
    f_{k,s}\!\bigl(
    \phi^{\ell-1}_k;\,
    \mathbf{x}_c,\,
    \mathbf{x}_{u,k},\,
    \xi^{\ell}_k
    \bigr),
    \qquad
    \forall\, s\in\mathcal{O}_k,
    \label{eq:phase2_grad}
\end{equation}
where $\xi^{\ell}_k$ denotes a mini-batch sampled from $D_k$. The client then computes the optimal personalised descent weights $\boldsymbol{\mu}^*_{k,\ell}\in\mathbb{R}^{m_k}$ by solving
\begin{equation}
\begin{aligned}
\boldsymbol{\mu}^*_{k,\ell}
&=
\arg\min_{\boldsymbol{\mu}\in\mathcal{C}_k}
\left\|
\sum_{s\in\mathcal{O}_k}
\mu_s\,
\mathbf{g}^{\ell}_{k,s}
\right\|^2,\\
\mathcal{C}_k
&=
\Bigl\{
\boldsymbol{\mu}\in[0,1]^{m_k}
:\;
\sum_{s\in\mathcal{O}_k}\mu_s=1
\Bigr\}.
\end{aligned}
\label{eq:phase2_qp}
\end{equation}
Using these weights, the client updates the personalised model along the local descent direction:
\begin{equation}
    \phi^{\ell}_k
    =
    \phi^{\ell-1}_k
    -
    \eta_p
    \sum_{s\in\mathcal{O}_k}
    \mu^*_{k,\ell,s}\,
    \mathbf{g}^{\ell}_{k,s},
    \label{eq:phase2_update}
\end{equation}
where $\eta_p>0$ denotes the personalised learning rate. The client continues this process until convergence, i.e., when
\[
\left\|
\sum_{s\in\mathcal{O}_k}
\mu^*_{k,\ell,s}\,
\mathbf{g}^{\ell}_{k,s}
\right\|^2
\le
\epsilon_k,
\]
or until $\ell=\tau$. The final personalised model at client $k$ is
\begin{equation}
    \phi_k^*
    \triangleq
    \phi_k^{\mathrm{final}},
    \label{eq:phase2_output}
\end{equation}
where $\phi_k^{\mathrm{final}}$ denotes the last iterate obtained after convergence or completion of $\tau$ local iterations. The personalised model $\phi_k^*$ is initialised from the globally trained backbone $w^*$ and therefore leverages shared knowledge across clients while adapting to client-specific features and objectives. Table~\ref{tab:notation} summarises the key notation used throughout the FedCARE framework. The complete federated multi-objective training procedure for Stage~2 is presented in Algorithm~\ref{alg:stage2}.
\begin{algorithm}
\caption{Stage~1: Federated Multi-Objective Training}
\label{alg:stage1}
\begin{algorithmic}[1]

\Require
$\{D_k\}_{k=1}^{K}$, $S$, $\eta_l$, $\eta_g$, $\tau$, $T$, $\epsilon$
\Ensure $w^*$

\vspace{2pt}
\State \textbf{Server:} initialise and broadcast $w^0$ to all clients $k \in \mathcal{K}$

\vspace{2pt}
\For{$t = 1, 2, \ldots, T$}

    \vspace{2pt}
    \State \hspace{0.5em} \textit{// Client}

    \ForAll{$k \in \mathcal{K}$ \textbf{in parallel}}

        \State Set $w^{t,0}_{k,s} \leftarrow w^t$ for all $s \in \mathcal{O}_k$

        \For{$\ell = 1,2,\ldots,\tau$}

            \For{each objective $s \in \mathcal{O}_k$}

                \State $w^{t,\ell}_{k,s} \leftarrow
                w^{t,\ell-1}_{k,s}
                - \eta_l\nabla f_{k,s}
                (w^{t,\ell-1}_{k,s};\,\mathbf{x}_c,\,\xi^{t,\ell}_k)$

            \EndFor

        \EndFor

        \For{each objective $s \in \mathcal{O}_k$}

            \State $\boldsymbol{\Delta}^t_{k,s}
            \leftarrow
            \sum_{\ell=1}^{\tau}
            \nabla f_{k,s}
            (w^{t,\ell-1}_{k,s};\,\mathbf{x}_c,\,\xi^{t,\ell}_k)$

        \EndFor

        \State Send $\{\boldsymbol{\Delta}^t_{k,s}\}_{s \in \mathcal{O}_k}$ to server

    \EndFor

    \vspace{2pt}
    \State \hspace{0.5em} \textit{// Server}

    \For{$s = 1,2,\ldots,S$}

        \State $\boldsymbol{\Delta}^t_s
        \leftarrow
        \sum_{k \in \mathcal{K}_s}
        p_{k,s}\,
        \boldsymbol{\Delta}^t_{k,s}$

    \EndFor

    \State $\boldsymbol{\lambda}^*_t
    \leftarrow
    \arg\min_{\boldsymbol{\lambda}\in\mathcal C}
    \bigl\|
    \sum_{s=1}^{S}
    \lambda_s\boldsymbol{\Delta}_s^t
    \bigr\|^2$

    \State $\mathbf d^t
    \leftarrow
    \sum_{s=1}^{S}
    \lambda_{t,s}^*
    \boldsymbol{\Delta}_s^t$

    \State $w^{t+1}\leftarrow w^t-\eta_g\mathbf d^t$

    \If{$\|\mathbf d^t\|^2\le\epsilon$}

        \State \textbf{break}

    \EndIf

    \State Broadcast $w^{t+1}$ to all clients

\EndFor

\State $w^*\leftarrow w^T$

\end{algorithmic}
\end{algorithm}

\begin{algorithm}
\caption{Stage~2: Multi-Objective Personalised Fine-Tuning}
\label{alg:stage2}
\begin{algorithmic}[1]

\Require $w^*$, $\{D_k\}_{k=1}^{K}$, $\eta_p$, $\tau$, $\epsilon_k$
\Ensure $\phi_k^*$ for each client $k \in \mathcal{K}$

\vspace{2pt}
\ForAll{$k \in \mathcal{K}$}

    \State $\phi_k^0 \leftarrow w^*$

    \For{$\ell = 1,2,\ldots,\tau$}

        \For{each objective $s \in \mathcal{O}_k$}

            \State $\mathbf{g}_{k,s}^{\ell}
            \leftarrow
            \nabla_{\phi_k}
            f_{k,s}
            (\phi_k^{\ell-1};\,
            \mathbf{x}_c,\,
            \mathbf{x}_{u,k},\,
            \xi_k^{\ell})$

        \EndFor

        \State $\boldsymbol{\mu}_{k,\ell}^{*}
        \leftarrow
        \arg\min_{\boldsymbol{\mu}\in\mathcal{C}_k}
        \left\|
        \sum_{s\in\mathcal{O}_k}
        \mu_s\,
        \mathbf{g}_{k,s}^{\ell}
        \right\|^2$

        \State $\phi_k^{\ell}
        \leftarrow
        \phi_k^{\ell-1}
        -
        \eta_p
        \sum_{s\in\mathcal{O}_k}
        \mu_{k,\ell,s}^{*}\,
        \mathbf{g}_{k,s}^{\ell}$

        \If{$\left\|
        \sum_{s\in\mathcal{O}_k}
        \mu_{k,\ell,s}^{*}\,
        \mathbf{g}_{k,s}^{\ell}
        \right\|^2
        \le
        \epsilon_k$}

            \State \textbf{break}

        \EndIf

    \EndFor

    \State $\phi_k^{*} \leftarrow \phi_k^{\ell}$

\EndFor

\end{algorithmic}
\end{algorithm}

\begin{table}[t]
\centering
\caption{Key notation used in the FedCARE framework.}
\label{tab:notation}
\renewcommand{\arraystretch}{1.15}
\begin{tabular}{ll}
\toprule
\textbf{Notation} & \textbf{Description} \\
\midrule
$K$ & Number of participating clients. \\
$S$ & Total number of distinct objectives. \\
$\mathcal{K}$ & Set of participating clients. \\
$\mathcal{O}_k$ & Set of objectives at client $k$. \\
$\mathcal{K}_s$ & Set of clients participating in objective $s$. \\
$m_k$ & Number of objectives at client $k$. \\
$\mathcal{X}_c$ & Common feature space shared by all clients. \\
$\mathcal{X}_{u,k}$ & Client-specific uncommon feature space. \\
$D_k$ & Local dataset of client $k$. \\
$f_{k,s}$ & Local objective function for objective $s$ at client $k$. \\
$w^t$ & Global backbone model at communication round $t$. \\
$w^*$ & Converged global backbone model. \\
$\phi_k^*$ & Personalised model of client $k$. \\
$\boldsymbol{\Delta}_{k,s}$ & Local gradient update for objective $s$. \\
$\boldsymbol{\Delta}_s$ & Aggregated global gradient update for objective $s$. \\
$\mathbf{g}_{k,s}$ & Local objective gradient during personalization. \\
$\boldsymbol{\lambda}$ & Global objective weight vector from the server QP. \\
$\boldsymbol{\mu}$ & Personalised objective weight vector from the client QP. \\
$\eta_l,\eta_g,\eta_p$ & Local, global, and personalised learning rates. \\
$\tau$ & Number of local training iterations. \\
$T$ & Number of communication rounds. \\
\bottomrule
\end{tabular}
\end{table}

\section{Performance Evaluation}
In this section, we evaluate the performance of the proposed FedCARE framework through experiments and compare it with standard FL methods. Specifically, Section V-A presents the experimental setup and datasets used. Section V-B presents a comparative analysis of FedCARE with baseline algorithms using AUROC and MAE. Section V-C evaluates the impact of the number of clients by varying it from 5 to 30 while keeping other parameters constant. In Section V-D, we perform an ablation analysis to evaluate personalised fine-tuning under feature disparity. In Section V-E, we evaluate performance per objective on both datasets.
\subsection{Experimental Setup}
We performed all experiments using the Flower simulation environment \cite{beutel2020flower} and deployed it on the MRC, comprising multiple Virtual Machines (VMs). One VM acts as the central server, while the remaining VMs represent independent healthcare organisations acting as federated clients. The framework is implemented in Python 3.10 with PyTorch for neural network training. The Pareto weight computation is solved using the Sequential Least Squares Programming (SLSQP) solver from SciPy. The central server is running on a VM with 4 vCPUs, 16 GB RAM, and 30 GB storage, and is configured with Ubuntu 22.04 LTS. The client nodes are deployed on VMs with 8 vCPUs and 32 GB RAM. All instances communicate over an internal network using secure SSH authentication. All models are optimised using local learning rate of $\eta_l = 0.001$ and a global learning rate of $\eta_g = 0.01$. The model is trained for $100$ federated rounds or until convergence, and each client performs $\tau = 5$ local updates per communication round.

\subsubsection{Datasets}
\begin{itemize}
\item \textbf{Diabetes 130-US Hospitals} \cite{strack2014impact}: This dataset contains 101,766 records with 50 features collected from 130 US hospitals over a period of 10 years. We consider three personalised tasks: early discharge risk prediction, readmission prediction, and length-of-stay prediction. We divide the feature space into 20 common features, and clients with tasks 1 and 2 each have 9 uncommon features, while clients with task 3 have 8 uncommon features and removed five low-variance features. 
    
\item \textbf{MIMIC-III} \cite{johnson2016mimic}: This dataset contains 21,139 ICU records with 49 features. We consider three tasks: mortality, readmission, and length-of-stay prediction. We partition the feature space into 19 common features shared across all clients, while each client uses 9 uncommon features specific to its assigned task. Additionally, we include 3 output features corresponding to the prediction tasks.
\end{itemize}
\subsubsection{Model Architectures}
\begin{itemize}
\item \textbf{Diabetes 130-US Hospitals} The shared backbone takes the common features $\mathbf{x}_c \in \mathbb{R}^{20}$ as input and consists of two fully connected layers with 128 units each and ReLU activation. Batch Normalisation and Dropout ($p=0.3$) are applied after each layer. A residual connection is used between the two layers to improve learning. The output is then reduced to a 64-dimensional representation using a linear layer followed by Layer Normalisation. This backbone is shared across clients and each client has its own local head, which takes the concatenation of the shared representation $\mathbf{h}_k \in \mathbb{R}^{64}$ and its private features. For tasks 1 and 2 (early discharge and readmission), the input size is 73 and the head has two layers (64 and 32 units with ReLU), followed by Dropout and a Sigmoid output for classification. For task 3 (length of stay), the input size is 72, and the same structure is used with a Softplus output for regression, trained on log-transformed targets. In baseline methods, all features are combined and zero-padded to size 29, and a single FullMLP backbone is used without separating shared and private features.
    
\item \textbf{MIMIC-III} A similar model design is used, with changes to the input size. The shared backbone takes common features $\mathbf{x}_c \in \mathbb{R}^{19}$ and uses two fully connected layers with 128 units, ReLU activation, Batch Normalisation, and Dropout ($p=0.3$). A residual connection is included between the layers. The output is reduced to a 64-dimensional representation using a linear layer and Layer Normalisation. This backbone is shared across clients. Each client has a local head that takes a 73-dimensional input formed by combining the shared representation and 9 private features. The head has two layers (64 and 32 units with ReLU and Dropout). The output layer depends on the task: Sigmoid is used for mortality and readmission classification, and Softplus is used for length-of-stay regression. Classification tasks use weighted binary cross-entropy to handle class imbalance, and regression uses Huber loss on log-transformed targets. For baseline methods, all features are zero-padded to size 28 and processed using a single FullMLP model without personalization.

\end{itemize}

\subsection{Comparison to Baselines}
We compared the performance of FedCARE against standard federated aggregation methods, including FedAvg and FedProx, multi-objective federated optimization methods FMGDA and FMSGDA, and a personalised healthcare FL framework PAFNet. All baseline methods were evaluated under identical experimental settings. The comparative performance of all methods is presented in Table \ref{tab:auroc_mae_comparison}.
\subsubsection{Classification Performance} 
\begin{figure*}[htbp]
    \centering
    \begin{subfigure}[b]{0.48\textwidth}
        \centering
        \includegraphics[width=\linewidth]{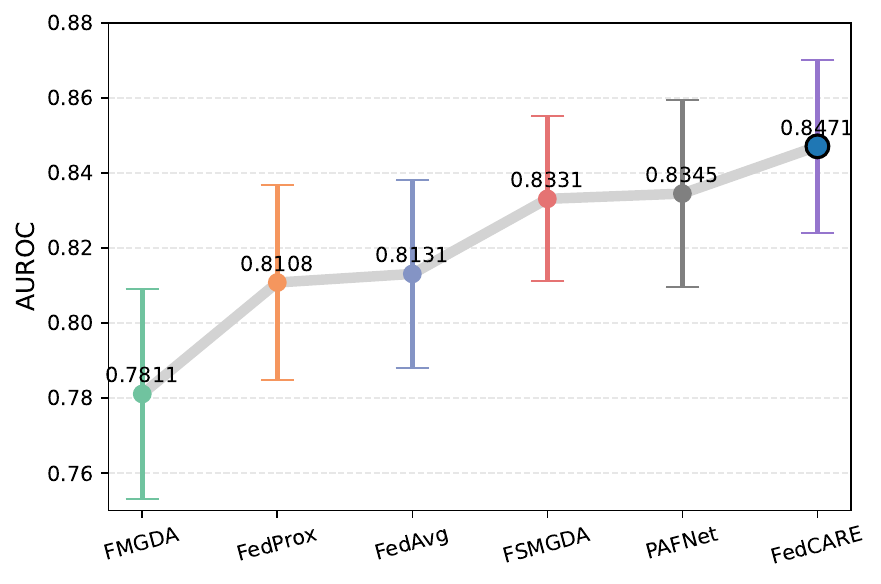}
        \caption{MIMIC-III}
        \label{fig:3a}
    \end{subfigure}
    \hfill
    \begin{subfigure}[b]{0.48\textwidth}
        \centering
        \includegraphics[width=\linewidth]{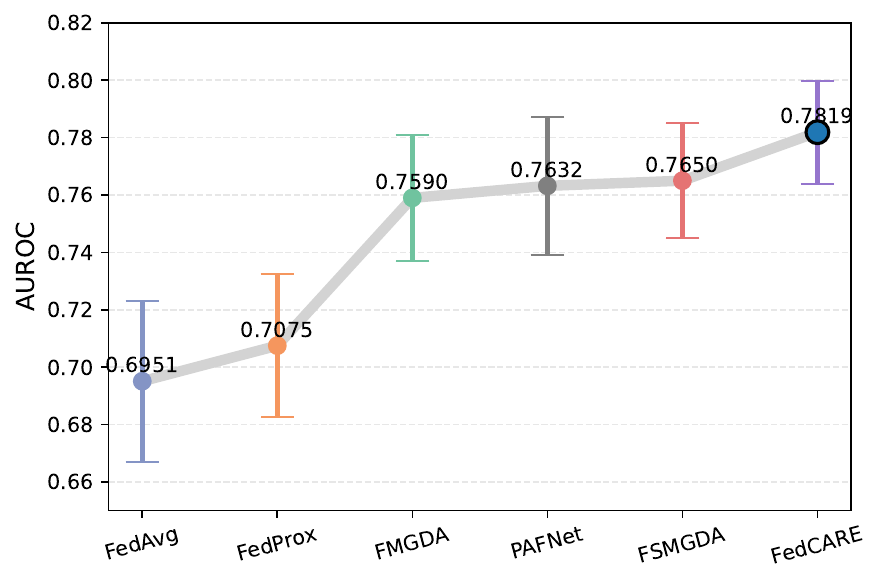}
        \caption{Diabetes 130-US Hospitals}
        \label{fig:3b}
    \end{subfigure}
    \caption{AUROC comparison of FedCARE and baseline methods on (a) MIMIC-III and (b) Diabetes 130-US Hospitals datasets.}
    \label{fig:combined_auroc}
\end{figure*}

Figure \ref{fig:3a} presents a comparison of AUROC of FedCARE and baseline methods on the MIMIC-III dataset. Among the standard federated aggregation methods, FedAvg and FedProx achieved AUROC values of $0.8131 \pm 0.025$ and $0.8108 \pm 0.026$, respectively. FMGDA obtained the lowest AUROC of $0.7811 \pm 0.028$, indicating that optimizing multiple conflicting objectives without personalization can adversely affect predictive performance. FSMGDA improved the AUROC to $0.8331 \pm 0.022$, demonstrating the benefits of stochastic multi-objective optimization. Similarly, PAFNet achieved an AUROC of $0.8345 \pm 0.025$, highlighting the effectiveness of personalised adaptation in healthcare FL environments. FedCARE achieved the highest AUROC of $0.8471 \pm 0.023$, outperforming FedAvg, FedProx, FMGDA, FSMGDA, and PAFNet by 4.18\%, 4.48\%, 8.45\%, 1.68\%, and 1.51\%, respectively. Although FSMGDA exhibited slightly lower variability, FedCARE consistently outperforms while providing personalisation. FedCARE achieved an average AUROC improvement of 4.06\% over all baseline methods, demonstrating its effectiveness in simultaneously addressing objective heterogeneity and feature-space heterogeneity in federated healthcare environments.\\
Figure \ref{fig:3b} presents the AUROC comparison on the Diabetes 130-US Hospitals dataset. FedAvg and FedProx achieved AUROC of $0.6951 \pm 0.028$ and $0.7075 \pm 0.025$, respectively, indicating limited robustness under heterogeneous healthcare objectives and feature distributions. The multi-objective approaches FMGDA and FSMGDA improved performance, achieving AUROC of $0.7590 \pm 0.022$ and $0.7650 \pm 0.020$, respectively. PAFNet achieved an AUROC of $0.7632 \pm 0.024$, comparable to that of $0.7632 \pm 0.024$, demonstrating the benefits of personalised learning in healthcare federated environments. FedCARE consistently achieved the best performance, with an AUROC of $0.7819 \pm 0.018$, and exhibited the smallest variation among all methods. Compared with FedAvg, FedProx, FMGDA, FSMGDA, and PAFNet, FedCARE improved AUROC by 12.49\%, 10.52\%, 3.02\%, 2.21\%, and 2.45\%, respectively. FedCARE achieved an average AUROC improvement of 6.14\% over all baseline methods.

\subsubsection{Regression Performance}
\begin{figure}[h!]
    \centering
    \includegraphics[width=\columnwidth]{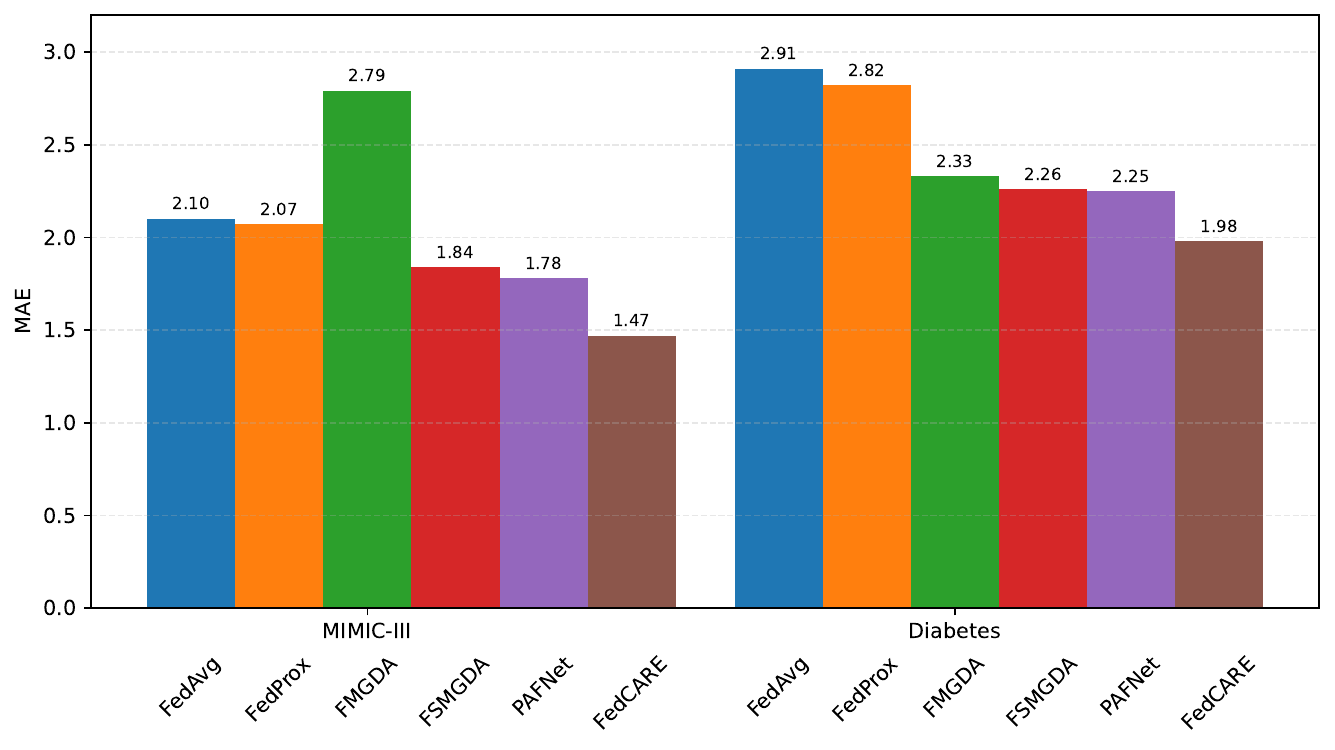}  
    \caption[Comparison of MAE across datasets]{MAE comparison of baseline algorithms with FedCARE across the MIMIC-III and Diabetes 130-US Hospitals datasets.}
    \label{fig:5}
\end{figure}

Figure~\ref{fig:5} shows the MAE results for the LOS regression task on both datasets, where lower values indicate better performance. FedCARE achieves the lowest MAE on both datasets, with 1.47 days on MIMIC-III and 1.98 days on Diabetes 130-US Hospitals. On MIMIC-III, FMGDA records the highest MAE of 2.79 days, performing less than FedAvg (2.10 days) and FedProx (2.07 days). FSMGDA improves the MAE to 1.84 days, while PAFNet further reduces it to 1.78 days. FedCARE achieves the lowest MAE of 1.47 days and reduces MAE by 28.77\% compared with all baseline methods.  Similar trends are observed on the Diabetes 130-US Hospitals dataset. FedAvg and FedProx achieve MAE values of 2.91 and 2.82 days, respectively, while FMGDA, FSMGDA, and PAFNet reduce the error to 2.33, 2.26, and 2.25 days. FedCARE achieves the lowest MAE of 1.98 days and reduces MAE by 20.23\% compared with all baselines. This demonstrates the effectiveness of FedCARE in handling heterogeneous objectives and feature spaces for accurate LOS prediction.

\begin{table}[h!]
\centering
\caption{AUROC and MAE comparison of five federated learning algorithms on MIMIC-III and Diabetes datasets.}
\label{tab:auroc_mae_comparison}
\begin{tabular}{l l c c}
\toprule
Dataset & Algorithm & AUROC & MAE \\
\midrule
\multirow{5}{*}{MIMIC-III} 
       & FedAvg    & 0.8131 & 2.10 \\
       & FedProx   & 0.8108 & 2.07 \\
       & FMGDA     & 0.7811 & 2.79 \\
       & FSMGDA    & 0.8331 & 1.84 \\
       & PAFNet    & 0.8345 & 1.78 \\
       & FedCARE   & 0.8471 & 1.47 \\
\midrule
\multirow{5}{*}{Diabetes 130-US Hospitals} 
       & FedAvg    & 0.6951 & 2.91 \\
       & FedProx   & 0.7075 & 2.82 \\
       & FMGDA     & 0.7592 & 2.33 \\
       & FSMGDA    & 0.7653 & 2.26 \\
       & PAFNet    & 0.7632 & 2.25 \\
       & FedCARE   & 0.7819 & 1.98 \\
\bottomrule
\end{tabular}
\end{table}

\subsection{Effect of Number of Clients}
\begin{figure*}[t]
\centering
\begin{subfigure}{0.48\textwidth}
    \centering
    \includegraphics[width=\linewidth]{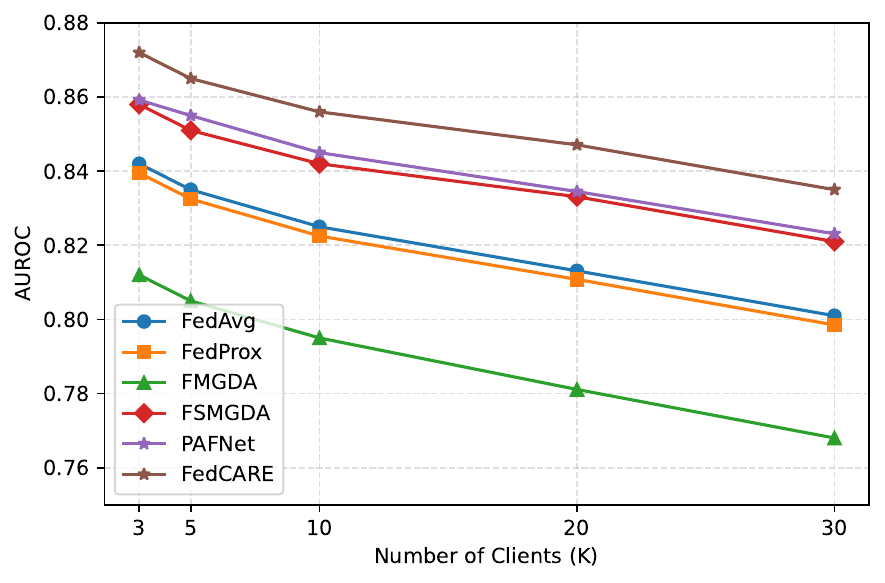}
    \caption{MIMIC-III dataset}
    \label{fig:scalability_mimic}
\end{subfigure}
\hfill
\begin{subfigure}{0.48\textwidth}
    \centering
    \includegraphics[width=\linewidth]{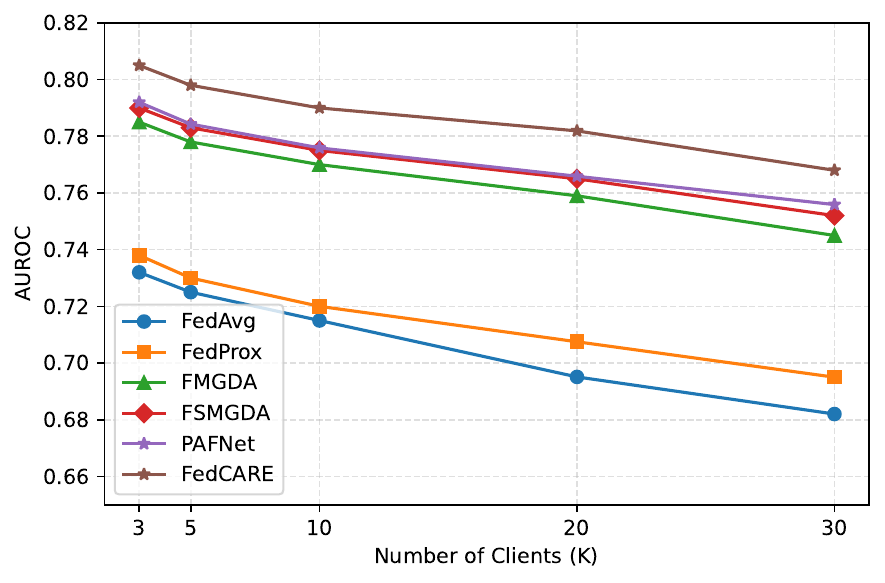}
    \caption{Diabetes 130-US Hospitals dataset}
    \label{fig:scalability_diabetes}
\end{subfigure}

\caption{Average AUROC across different client configurations.}
\label{fig:scalability}
\end{figure*}

To examine the behaviour of FedCARE to increasing federation size, we experimented with $K \in \{3, 5, 10, 20, 30\}$ clients. At each configuration, clinical objectives are drawn uniformly at random from the objective set $\{\text{early discharge risk},\ \text{readmission},\ \text{length-of-stay prediction}\}$ for the Diabetes 130-US Hospitals dataset and $\{\text{mortality},\ \text{readmission},\ \text{length-of-stay prediction}\}$ for the MIMIC-III dataset, such that multiple clients may share the same objective. We report the average AUROC across different client configurations, as illustrated in Figure~\ref{fig:scalability}(a) and (b) for the MIMIC-III and Diabetes datasets, respectively.  Two consistent trends are observed across both datasets and all algorithms. First, average accuracy decreases as the number of clients increases. This behaviour is attributed to two factors: (i) increased client introduces greater gradient divergence during federated aggregation; and (ii) the per-client data partition size decreases as the cohort is divided among more participants. Second, FedCARE consistently achieves higher average accuracy than all baseline algorithms across every client configuration.  

\subsection{Impact of Feature Disparity Ablation Analysis}
\begin{figure*}[t]
    \centering
    
    \begin{subfigure}{0.48\textwidth}
        \centering
        \includegraphics[width=\linewidth]{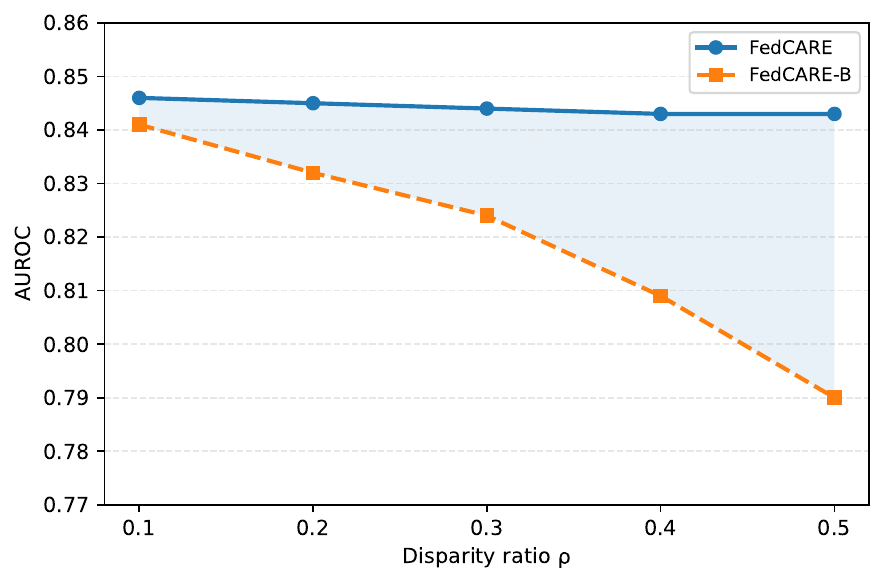}
        \caption{MIMIC-III Dataset}
        \label{fig:mimic_auroc}
    \end{subfigure}
    \hfill
    \begin{subfigure}{0.48\textwidth}
        \centering
        \includegraphics[width=\linewidth]{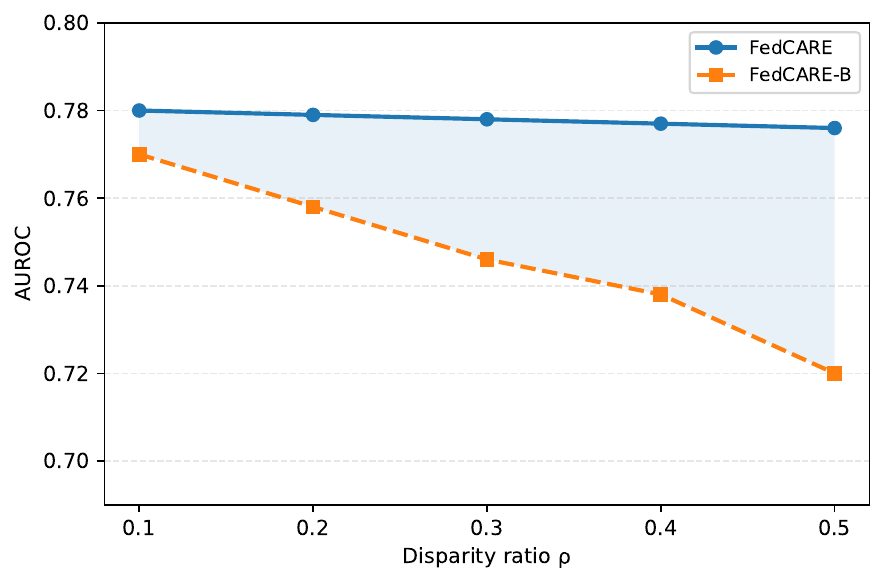}
        \caption{Diabetes 130-US Dataset}
        \label{fig:diabetes_auroc}
    \end{subfigure}
    
    \caption{Impact of feature disparity ($\rho$) on AUROC performance across datasets.}
    \label{fig:auroc_combined}
\end{figure*}

\begin{figure}[h]
    \centering
    \includegraphics[width=\columnwidth]{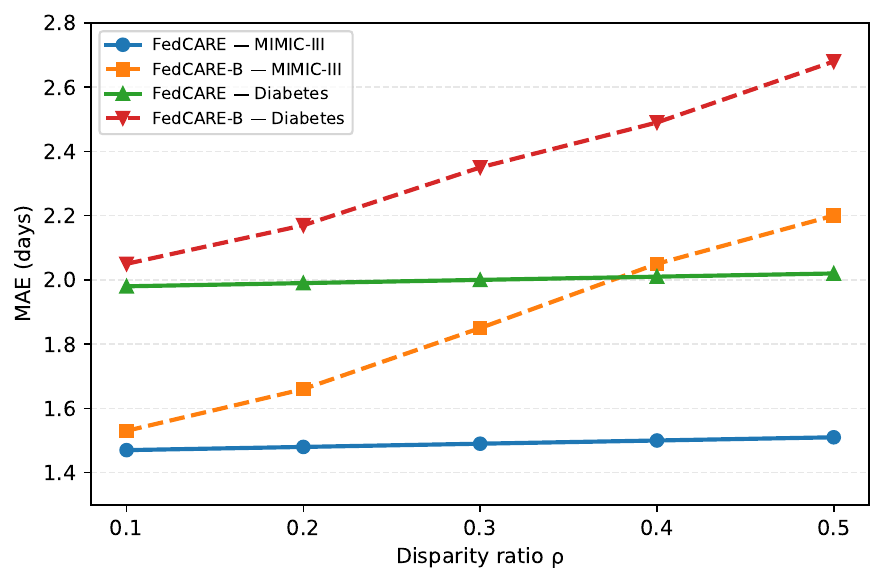}
    \caption{Impact of feature disparity ($\rho$) on MAE (days) for both datasets.}
    \label{fig:mae}
\end{figure}

To evaluate personalised fine-tuning under feature disparity, we conduct an ablation study comparing the FedCARE framework with a backbone-only variant, denoted FedCARE-B. In this variant, Stage~2 is omitted and inference is performed directly using the globally trained backbone $w^*$ on common features alone.
This comparison quantifies the benefit of the personal head $\phi_k^*$ trained on private uncommon features $\mathbf{x}_{u,k}$ under varying degrees of feature disparity. Feature disparity is controlled by varying the proportion of uncommon features relative to the total feature set. We define the disparity ratio as: 
\[
\rho = \frac{d_u}{d_c + d_u},
\]
where $d_c$ denotes the number of common features and $d_u$ the number of uncommon features per client. We evaluate five disparity settings $\rho \in \{0.1,\, 0.2,\, 0.3,\, 0.4,\, 0.5\}$, corresponding to scenarios where uncommon features constitute between $10\%$ and $50\%$ of the total available feature space. At $\rho = 0.1$, clients have nearly identical feature spaces with minimal private information, whereas at $\rho = 0.5$, half of each client's features are institution-specific and unavailable to other participants. All other hyperparameters are held constant across settings. Results are averaged over ten independent runs on both the MIMIC-III and Diabetes 130-US Hospitals datasets and are reported in terms of average AUROC across binary classification tasks and MAE for the length-of-stay regression task. The AUROC results are illustrated in Figure~\ref{fig:auroc_combined}(a) for the MIMIC-III dataset and Figure~\ref {fig:auroc_combined}(b) for the Diabetes 130-US Hospitals dataset; the MAE results for both datasets are presented in Figure~\ref {fig:mae}.

Two consistent trends are observed across both datasets. First, as the disparity ratio $\rho$ increases, the performance of FedCARE-B degrades gradually. At $\rho = 0.1$, FedCARE-B achieves an AUROC of $0.841$ on MIMIC-III and $0.771$ on Diabetes, which is close to the full FedCARE performance, as the backbone trained on common features captures nearly all available clinical signal when the uncommon feature set is small. However, as $\rho$ increases to $0.5$, FedCARE-B degrades substantially to AUROC of $0.794$ on MIMIC-III and $0.718$ on Diabetes, a drop of $5.6\%$ and $6.9\%$ respectively, as an increasing proportion of clinically informative private features is excluded from the global model. This degradation shows that the backbone alone is insufficient when clients possess institution-specific feature sets. Second, FedCARE consistently maintains strong performance across all disparity settings on both datasets. On MIMIC-III, FedCARE achieves AUROC values ranging from $0.846$ at $\rho = 0.1$ to $0.843$ at $\rho = 0.5$, a negligible degradation of $0.4\%$ across the full disparity range. On Diabetes, FedCARE achieves AUROC values between $0.780$ and $0.776$, a degradation of less than $0.5\%$. The near-constant performance of FedCARE across all $\rho$ values demonstrates that the personalised fine-tuning stage effectively absorbs the information contained in private uncommon features, maintaining predictive performance even as feature disparity increases substantially. A similar trend is observed for the MAE regression metric: FedCARE-B MAE increases from $1.52$ to $2.18$ days on MIMIC-III as $\rho$ grows from $0.1$ to $0.5$, while FedCARE maintains a stable MAE between $1.47$ and $1.51$ days across the same range. As the value of $\rho$ increases, the difference in performance between FedCARE and FedCARE-B also increases. At $\rho = 0.5$, the gap is highest, with AUROC differences of 0.049 on MIMIC-III and 0.058 on the Diabetes dataset. This shows that the personal head becomes more important when clients have more private features.
\subsection{Performance per Objective}
To evaluate how the two-stage training performs on each distinct clinical objective individually, we report the per-objective test results for each participating client across both datasets. Since all algorithms are evaluated under the same multi-objective federated setting, the key distinction between methods lies in the server-side aggregation strategy and the use of private features: FedAvg and FedProx apply uniform gradient averaging, FMGDA and FSMGDA apply Pareto-optimal weighting, PAFNet applies personalised client-specific heads without explicit Pareto-optimal aggregation, and FedCARE combines Pareto-optimal aggregation with personalised fine-tuning on private uncommon features.

\subsubsection{MIMIC-III Dataset}

Table~\ref{tab:per_obj_mimic} shows the performance of all methods on the three MIMIC-III clinical tasks: in-hospital mortality prediction, ICU length-of-stay (LOS) regression, and 30-day readmission prediction. FedCARE achieves the best performance on all three objectives. For mortality prediction, FedCARE achieves an AUROC of 0.870, compared with 0.855 for PAFNet, 0.851 for FSMGDA, and 0.831 for FedAvg, resulting in improvements of 1.8\%, 2.2\%, and 4.7\%, respectively. For the LOS regression task, FedCARE achieves the lowest MAE of 1.47 days, improving over PAFNet (1.78 days), FSMGDA (1.84 days), and FedAvg (2.10 days) by 17.4\%, 20.1\%, and 30.0\%, respectively. FMGDA performs worst on LOS (2.79 days), even lower than FedAvg (2.10 days), because it is sensitive to the different scales of regression and classification losses, leading to poor objective balancing. FSMGDA improves the regression performance by using stochastic gradient estimation, while PAFNet further reduces the MAE through personalised local learning. By combining Pareto-optimal aggregation with Stage~2 personalised fine-tuning, FedCARE achieves the best overall regression performance. For the readmission prediction task, FedCARE also achieves the highest AUROC of 0.824, followed by PAFNet (0.814) and FSMGDA (0.815). The improvement is smaller than for the other tasks, as expected, because 30-day readmission prediction is inherently more challenging with only in-hospital EHR data.

\begin{table}[h]
\centering
\caption{Per-objective test performance on MIMIC-III.}
\label{tab:per_obj_mimic}
\renewcommand{\arraystretch}{1.1}
\resizebox{\columnwidth}{!}{
\begin{tabular}{llcccccc}
\toprule
\textbf{Objective} & \textbf{Metric} & \textbf{FedAvg} & \textbf{FedProx} & \textbf{FMGDA} & \textbf{FSMGDA} & \textbf{PAFNet} & \textbf{FedCARE} \\
\midrule
Mortality     & AUROC   & 0.831 & 0.829 & 0.801 & 0.851 & 0.855 & \textbf{0.870} \\
LOS           & MAE (d) & 2.10  & 2.07  & 2.79  & 1.84  & 1.78  & \textbf{1.47}  \\
Readmission   & AUROC   & 0.795 & 0.793 & 0.761 & 0.815 & 0.814 & \textbf{0.824} \\
\bottomrule
\end{tabular}
}
\end{table}

\subsubsection{Diabetes 130-US Hospitals Dataset}

Table~\ref{tab:per_obj_diabetes} reports the per-objective results on the three Diabetes tasks: early discharge risk prediction, 30-day readmission prediction, and length-of-stay regression. FedCARE achieves the best performance on all three objectives. For early discharge prediction, FedCARE achieves an AUROC of 0.801, compared with 0.775 for PAFNet and 0.779 for FSMGDA, corresponding to improvements of 3.4\% and 2.8\%, respectively. This improvement is mainly because FedCARE uses private client features, such as admission type, medical specialty, and discharge disposition, which provide important information for predicting early discharge. Although PAFNet uses personalised learning, FSMGDA performs slightly better on this task because its Pareto-based aggregation handles multiple objectives more effectively on the Diabetes dataset. A similar trend is observed for readmission prediction, where FedCARE achieves an AUROC of 0.771, compared with 0.754 for FSMGDA, 0.751 for PAFNet, and 0.681 for FedAvg. For the LOS regression task, FedCARE achieves the lowest MAE of 1.98 days, improving over PAFNet (2.25 days), FSMGDA (2.26 days), and FedAvg (2.91 days) by 12.0\%, 12.4\%, and 32.0\%, respectively. Unlike the MIMIC-III dataset, FMGDA performs much better than FedAvg on all Diabetes tasks. This is because the objective gradients are more balanced, allowing Pareto-based optimisation to find better solutions. Overall, these results show that combining Pareto-optimal aggregation with personalised fine-tuning enables FedCARE to achieve the best performance across all Diabetes prediction tasks.

\begin{table}[h]
\centering
\caption{Per-objective test performance on Diabetes 130-US Hospitals.}
\label{tab:per_obj_diabetes}
\renewcommand{\arraystretch}{1.1}
\resizebox{\columnwidth}{!}{
\begin{tabular}{llcccccc}
\toprule
\textbf{Objective} & \textbf{Metric} & \textbf{FedAvg} & \textbf{FedProx} & \textbf{FMGDA} & \textbf{FSMGDA} & \textbf{PAFNet} & \textbf{FedCARE} \\
\midrule
Early discharge  & AUROC   & 0.712 & 0.724 & 0.773 & 0.779 & 0.775 & \textbf{0.801} \\
Readmission      & AUROC   & 0.681 & 0.694 & 0.748 & 0.754 & 0.751 & \textbf{0.771} \\
LOS              & MAE (d) & 2.91  & 2.82  & 2.33  & 2.26  & 2.25  & \textbf{1.98}  \\
\bottomrule
\end{tabular}
}
\end{table}

Across all objectives and both datasets, FedCARE consistently achieves the best performance. Uniform averaging (FedAvg, FedProx) suffers from gradient conflict under heterogeneous objectives. Pareto-optimal aggregation (FMGDA, FSMGDA) mitigates this conflict, with FSMGDA providing more reliable gains. PAFNet's client-level personalisation is competitive on tasks with strong private-feature signal but lacks in balancing multi-objective, and FedCARE further improves every objective through Stage~2 personalisation on private institution-specific features combined with Pareto-optimal aggregation, with the largest gains on objectives whose predictive signal is concentrated in private uncommon features.

\section{Conclusions and Future Work}

We proposed \textbf{FedCARE}, a multi-objective personalised FL framework for healthcare environments with objective heterogeneity and feature space disparity. FedCARE addresses settings where institutions optimise different and potentially conflicting clinical objectives while sharing only partially overlapping feature spaces. Its two-stage training strategy first learns a Pareto-stationary global backbone from common features and then enables each client to personalise the model using private features and local objectives, without additional communication during personalisation. Experiments on MIMIC-III and Diabetes 130-US Hospitals show that FedCARE consistently outperforms standard FL, multi-objective FL, and personalised FL baselines in terms of AUROC and MAE. These results demonstrate the benefit of jointly modelling conflicting objectives and institution-specific feature spaces in federated healthcare systems. The cloud-based client-server deployment further highlights the practical feasibility of FedCARE for distributed healthcare service computing. Future work will investigate client clustering for collaborative personalisation, robustness to dynamic client participation and stragglers, and stronger privacy-preserving mechanisms such as secure aggregation and differential privacy.

\bibliographystyle{IEEEtran}
\bibliography{references}

\end{document}